\documentclass[10pt,twocolumn,letterpaper]{article}

\usepackage{cvpr}              

\usepackage{graphicx}
\usepackage{amsmath}
\usepackage{amssymb}
\usepackage{booktabs}
\usepackage{algorithm}
\usepackage{algpseudocode}

\usepackage[pagebackref,breaklinks,colorlinks]{hyperref}

\usepackage[capitalize]{cleveref}
\crefname{section}{Sec.}{Secs.}
\Crefname{section}{Section}{Sections}
\Crefname{table}{Table}{Tables}
\crefname{table}{Tab.}{Tabs.}

\def\cvprPaperID{37} 
\def\confName{CVPR}
\def\confYear{2023}

\begin{document}

\title{SigSegment: A Signal-Based Segmentation Algorithm for Identifying Anomalous Driving Behaviours in Naturalistic Driving Videos}

\author{\thanks{Corresponding author and primary author} Kelvin Kwakye \& Younho Seong\\
Department of Systems and Industrial Engineering\\
North Carolina A\&T State University\\
\and
Armstrong Aboah \\
Department of Radiology\\
Northwestern University\\
\and
Sun Yi\\
Department of Systems and Industrial Engineering\\
North Carolina A\&T State University\\
}
\maketitle

\begin{abstract}
   In recent years, distracted driving has garnered considerable attention as it continues to pose a significant threat to public safety on the roads. This has increased the need for innovative solutions that can identify and eliminate distracted driving behavior before it results in fatal accidents. In this paper, we propose a Signal-Based anomaly detection algorithm that segments videos into anomalies and non-anomalies using a deep CNN-LSTM classifier to precisely estimate the start and end times of an anomalous driving event. In the phase of anomaly detection and analysis, driver pose background estimation, mask extraction, and signal activity spikes are utilized. A Deep CNN-LSTM classifier was applied to candidate anomalies to detect and classify final anomalies. The proposed method achieved an overlap score of \textbf{0.5424} and ranked \textbf{9th} on the public leader board in the AI City Challenge 2023, according to experimental validation results.
\end{abstract}

\section{Introduction}
\label{sec:intro}

In the United States, distracted driving is a serious threat to public safety, resulting in numerous fatalities each year\cite{national2021traffic}. Researchers have turned to naturalistic driving studies and computer vision techniques in order to identify and eliminate distracting driving behaviors\cite{fridman2019advanced},\cite{trivedi2007looking}. However, these methods face obstacles such as inadequate data labeling, poor data quality, and low resolution\cite{jegham2020vision,shoman2022region}.

Naturalistic driving studies or videos provide data on all aspects of the driver's actions, including drowsiness and distracted driving, making them a valuable tool for understanding driver behavior in real-world scenarios\cite{dingus2016driver}. Traditional machine learning approaches have been used in previous studies to present various driver distraction activity analyses and feature extraction methods\cite{frohlich2014will},\cite{lefevre2013learning}. For example, Braunagel et al. proposed using a Support Vector Machine (SVM) model to recognize driver activity using five contextual features: saccades, fixations, blink, head position, and rotation\cite{braunagel2015driver,aboah2023driver}. Similarly, Liang et al. used an SVM model and a logistic regression model to identify driver cognitive distractions in real-time using driver eye movement and vehicle dynamic data \cite{liang2007real}.
Also, in recent years, deep learning approaches have also been used in recent years for activity analysis and feature extraction in driver distraction detection. Vijayan., for example, proposed a deep convolutional neural network (CNN) architecture for detecting driver drowsiness based on features extracted from the driver's face, such as eye, mouth, and brow movements\cite{vijayan2019real,aboah2023ai,Aboah23AIC23}. Similarly, Omerustaoglu et al.\cite{omerustaoglu2020distracted} used a deep neural network (DNN) to detect driver distraction using features extracted from the driver's face, arms, and hands movements in detecting distracted and non-distracted states. Although these Driver distraction detection systems have shown promising results in identifying anomalous driving events using traditional machine learning approaches and deep learning such as activity analysis and feature extraction. However, due to the complexity and dynamic nature of driving behavior, more research is required to investigate the generalization and scalability of these approaches across a wide range of driving conditions and environments.

Due to the complexity and dynamic nature of driving behavior, identifying anomalous driving events in these videos is difficult. In response to this difficulty, we present SigSegment, a signal-based segmentation algorithm that employs pose estimation, signal generation, activity classification using deep learning techniques to identify anomalous driving events in natural driving videos. SigSegment uses a deep convolutional neural network (CNN) and long short-term memory (LSTM) architecture to classify anomalous driving events precisely by estimating the start and end times.

The remainder of the paper is structured as follows. The second section is a review of relevant literature. Section three contains the data and proposed methodology used for this study. The fourth section discusses the model development's results. Section five concludes with a summary of the findings, and recommendations for future research.

\subsection{Objective}

The objective of this study is to develop a classification model that accurately identifies distracted behavior activities executed by drivers within a given time frame.


\section{Related Works}
\label{sec:Work}

Driver distraction is a major concern in transportation safety. In recent years, researchers have used machine learning and deep learning techniques to classify driver distraction activities. In the literature, two main methods have been proposed: the single-frame-based method and the consecutive-sequence-based method\cite{pan2021driver,Aboa23AICITY23}.
The single-frame method analyzes a single frame of a video feed or image and applies machine learning algorithms, such as convolutional neural networks (CNNs), to classify the driver's behavior. While this method has shown high accuracy rates for certain types of distractions, such as cell phone use, it has limitations when it comes to capturing the temporal aspect of driver behavior. Several studies have been conducted to investigate the efficacy of the single-frame-based method in detecting different types of driver distraction using only a single image or video frame.
Lee et al.\cite{lee2018convolutional}, for example, proposed a method for detecting aggressive driving using CNN. Near-infrared (NIR) light and thermal camera sensors were used to capture the driver's face, and the CNN achieved a high accuracy rate of 99.95\%. Similarly, Guo et al. \cite{guo2019driver}  proposed a single-frame hybrid of convolutional neural network (CNN) and long short-term memory (LSTM)-based driver drowsiness detection system. CNN was trained to recognize drowsy facial features like drooping eyelids and yawning and achieved a high accuracy rate in classifying drowsy and alert states using only a single frame of the driver's face.
Liang et al. \cite{liang2018performance} proposed a real-time system for detecting driver cognitive distraction based on a single-frame Bayesian network in another study. The network was trained to model the relationship between the driver's eye movement and driving performance, and it detected cognitive distraction with an overall accuracy rate of 80.1\%. Qu et al. (2020) proposed a system for detecting driver phone use that was based on a single-frame CNN that analyzed the driver's hand and arm positions and achieved a 93.3\% accuracy rate. Furthermore, Zhang et al. (2019) proposed a system for detecting driver smoking based on a single-frame CNN that recognized the distinctive motion of a smoking hand and achieved 92.3\% accuracy\cite{aboah2021vision}.
Despite the single-frame-based method's demonstrated effectiveness in detecting various types of driver distraction. It is critical to recognize its limitations in capturing the temporal aspect of driver behavior, which the consecutive-sequence-based method may address more effectively by extracting temporal high-level features from sequential frames. 

The consecutive-sequence-based method, on the other hand, entails analyzing a continuous sequence of frames or video feed to identify patterns of behavior over time. This method, which typically employs recurrent neural networks (RNNs) and other time-series analysis techniques, has been demonstrated to be effective in identifying more complex types of driver distraction activities, such as touching the radio, eating or drinking while driving\cite{carvalho2017exploiting}. Sun et al. \cite{sun2020research}  proposed a method for detecting driver cognitive distraction using an RNN-based approach in one study. To detect cognitive distraction, the authors used the RNN's attention mechanism to focus on important regions of the driver's face, such as the eyes and mouth. The proposed method detected cognitive distraction with an accuracy rate of 90.64\%. Ed-Doughmi et al.\cite{ed2020real}  proposed a system for detecting driver phone use based on a consecutive-sequence-based approach in another study. To detect phone use, the system used a two-stream RNN to analyze both the driver's hand and face movements. The proposed method detected phone use with an accuracy rate of 92.1\%. Similarly, Bai et al.\cite{bai2021two} proposed a method for detecting driver drowsiness based on consecutive frames of the driver's face in their study. The authors achieved an accuracy rate of 93.4\% by using a two-stream Graph CNN to analyze the temporal changes in facial features associated with drowsiness, such as eye closure and head nodding\cite{boah2021mobile}.
Because of the rapid increase in the number of researchers and techniques, the accuracy of human driver distraction activity recognition has significantly improved. However, due to several issues such as inadequate labeling, poor data quality, low resolution, large intra-class variability, indistinct boundaries between classes, varying viewpoints, occlusions, appearance variations, influence of environmental factors, and recording settings, this task remains difficult\cite{ramanathan2014human,aboah2020smartphone}. These difficulties are exacerbated when dealing with naturalistic driving videos.

In this study, we propose we propose a Signal-Based anomaly detection algorithm that segments videos into anomalies and non-anomalies using a deep CNN-LSTM classifier to precisely estimate the start and end times of an anomalous
driving event.

\section{Method}
\label{sec:method}

This section describes the dataset and algorithm framework used in this study. Section 3.1, in particular, introduces the dataset used. The video pre-processing and segmentation based on the algorithm, the CNN frameworks, as well as the LSTM framework, are described in Section 3.2

\subsection{Dataset}
The dataset was provided by AI CITY CHALLENGE 2023. It includes 210 video clips captured by 35 drivers for a total of approximately 34 hours. The drivers performed 16 random tasks, such as talking on the phone, eating, and reaching back, which were recorded simultaneously by three cameras mounted in the car from various angles. Each driver performed the data collection tasks twice: once without any appearance block such as a hat or sunglasses, and once with such a block. This resulted in 6 videos per driver, with 3 synchronized without an appearance block and 3 with an appearance block.
The 34 hours of video footage was divided into three datasets, labeled A1, A2, and B, containing 25, 5, and 5 drivers, respectively. The A1 dataset contains manually annotated ground truth labels for the start time, end time, and type of distracted behavior. The dataset A2 contains no labels and Dataset B is for testing.
\subsection{The Proposed Framework}

\begin{figure*}[t!]
  \centering
  \includegraphics[width=17cm]{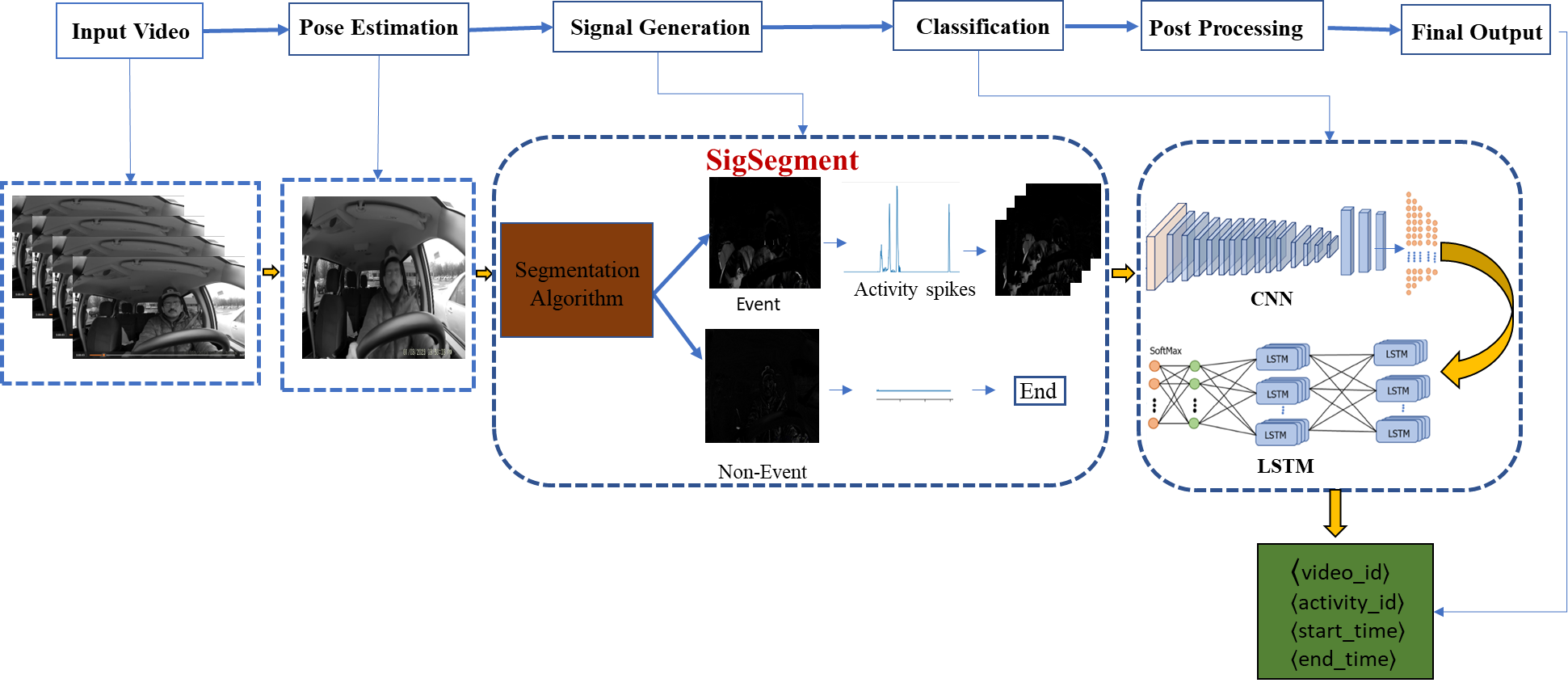}
  \caption{The Framework of the proposed Sigsegment CNN-LSTM model}
  \label{fig:example}
\end{figure*}

The proposed framework for the Naturalistic Driving Action Recognition consists of several steps, beginning with video data input, pose estimation, signal generation, activity classification and postprocessing as shown in  Figure \ref{fig:example}.

\vspace{0.1in}
\noindent\textbf{Posture Estimation}. Posture estimation is the process of determining the position and orientation of the driver's body. This information is important for identifying unusual driving behaviors, as changes in posture may indicate that the driver is becoming distracted or fatigued. To estimate the driver's posture from the video, we find the mean $\bar{X}$ of all frames in the video, where $X_i$ denotes the posture of the driver in the $i$-th frame (see Equation~\ref{eq:p1}). The assumption is that the normal driving posture $\bar{X}$ is the most frequent and probable driver activity in any driving section.

\begin{equation}
\label{eq:p1}
\bar{X} = \frac{1}{n} \sum_{i=1}^{n} X_i
\end{equation}

where $n$ is the total number of frames in the video.

 
\vspace{0.1in}
\noindent\textbf{Signal Generation}.  The estimated posture is subtracted from each frame in the video, yielding an image that is flattened and its pixel intensities summed as shown in Algorithm~\ref{alg:p2}. The resulting summation values serve as a quantitative measure of how different each frame is from the expected normal posture. Specifically, frames exhibiting close adherence to the expected posture will have small summation values, while those deviating from the expected posture will have large values. This usually results in a spike when plotted. Frames corresponding to the spike are considered to potentially indicate an anomaly.

\begin{algorithm}[H]
\caption{Anomaly Detection based on Posture Estimation}
\label{alg:p2}
\begin{algorithmic}[1]
\Procedure{DetectAnomalies}{$X_1, X_2, ..., X_n, \hat{X}, k$}
\State $A \gets \varnothing$ 
\For{$i \gets 1$ to $n$}
\State $Y_i \gets X_i - \hat{X}$ 
\State $V_i \gets \sum_{j=1}^p Y_{i,j}$ \Comment{Flatten and sum pixel intensities}
\EndFor
\State $m \gets \mathrm{median}(V_1, V_2, ..., V_n)$ \Comment{Compute median}
\State $S \gets \mathrm{std}(V_1, V_2, ..., V_n)$ \Comment{Compute standard deviation}
\State $\textit{thresh} \gets m + k \cdot S$ \Comment{Set anomaly detection threshold}
\For{$i \gets 1$ to $n$}
\If{$V_i > \textit{thresh}$}
\State $A \gets A \cup {X_i}$ \Comment{Add potential anomaly to set $A$}
\EndIf
\EndFor
\State \textbf{return} $A$ 
\EndProcedure
\end{algorithmic}
\end{algorithm}

\vspace{0.1in}
\noindent\textbf{Classification}. The classification model consists of two components: a convolutional neural network (CNN) to extract information from the image sequence, and a long short-term memory (LSTM) component to learn the temporal relationship between the frames in the input sequence. The results from the LSTM is passed through a fully connected neural network for the classification. 

\vspace{0.1in}
\noindent\textbf{Postprocessing}. The postprocessing step involves preparing the classified anomalies into the format required by the evaluation system.

\section{Results and Discussions}
\label{sec:Results}
The performance of our proposed model was assessed based on its ability to detect anomalies in driver distraction videos. Track 3 was evaluated based on the performance of model activity identification, as measured by the average activity overlap score, which is defined as follows. Given a ground-truth activity \textit{g} with start time gs and end time \textit{ge}, the closest predicted activity match is that predicted activity \textit{p} of the same class as g and highest overlap score \textit{os}, with the additional condition that start time \textit{ps} and end time \textit{pe} are in the range [gs – 10s, gs + 10s] and [ge – 10s, ge + 10s], respectively. The overlap between \textit{g} and \textit{p} is defined as the proportion of the time intersection to the time union of the two activities.

\begin{equation}
\label{eq:1}
o s(p, g)=\frac{\max (\min (g e, p e)-\max (g s, p s), 0)}{\max (g e, p e)-\min (g s, p s)}
\end{equation}

With an overlap score of \textbf{0.5424}, our proposed model was ranked \textbf{ninth} in the AI City Challenge 2023. The top-performing teams from the challenge's public leaderboard are shown in Table 1.
Our findings indicate that our proposed model can effectively detect driver distraction videos with anomalies, as evidenced by its F1score performance. Further enhancements can be made by experimenting with different evaluation metrics and optimizing the model's architecture.

\begin{table}[!ht]
    \centering
    \begin{tabular}{|l|l|l|l|}
    \hline
        Rank & TeamID & Team Name & Score \\ \hline
        1 & 209 & Meituan-loTCV & 0.7416 \\ \hline
        2 & 60 & JNU\_boat & 0.7041 \\ \hline
        3 & 49 & ctc-AI & 0.6723 \\ \hline
        4 & 118 & RW & 0.6245 \\ \hline
        5 & 8 & Purdue Digital Twin Lab & 0.5921 \\ \hline
        6 & 48 & BUPT-MCPRL & 0.5907 \\ \hline
        7 & 83 & DiveDeeper & 0.5881 \\ \hline
        8 & 217 & INTELLI\_LAB & 0.5426 \\ \hline
        \textbf{9} & \textbf{152} & \textbf{AILAB} & \textbf{0.5424} \\ \hline
        10 & 11 & AIMIZ & 0.5409 \\ \hline
    \end{tabular}
\end{table}

\section{Conclusion and Futureworks}
\label{sec:Conc}
In conclusion, this paper proposes a Signal-Based anomaly detection algorithm designed to detect  distracted driving behavior, which poses a significant threat to road safety. In the phase of anomaly detection and analysis, the proposed method employs driver pose background estimation, mask extraction, and signal activity spikes, which increases the accuracy of identifying anomalous driving events. Moreover, a Deep CNN-LSTM classifier was applied to candidate anomalies for accurate detection and analysis of final anomalies.
The experimental validation of the proposed method in the AI City Challenge 2023 demonstrated its efficacy, with an overlap score of 0.5424 and a ranking of ninth among other competitors. These results demonstrate the potential of applying deep learning techniques, in particular CNN-LSTM, to effectively detect and analyze anomalous driving behavior.
Future research could investigate the use of more sophisticated CNN-LSTM architectures or other deep learning models to enhance the detection and analysis of driving anomalies. In addition, the proposed method could be expanded to detect other types of driver distraction or incorporated into existing advanced driver assistance systems to further improve road safety.

{\small
\bibliographystyle{unsrt}
\bibliography{egbib}
}

\end{document}